# A machine-learning framework for daylight and visual comfort assessment in early design stages

Hanieh Nourkojouri[1], Zahra Sadat Zomorodian[1], Mohammad Tahsildoost[1], Zohreh Shaghaghian[2]
[1]Shahid Beheshti University, Tehran, Iran
[2]Texas A&M University, College Station, United States

## Abstract

This research is mainly focused on the assessment of machine learning algorithms in the prediction of daylight and visual comfort metrics in the early design stages. A dataset was primarily developed from 2880 simulations derived from Honeybee for Grasshopper. The simulations were done for a shoebox space with a one side window. The alternatives emerged from different physical features, including room dimensions, interior surfaces reflectance, window dimensions and orientations, number of windows, and shading states. 5 metrics were used for daylight evaluations, including UDI, sDA, mDA, ASE, and sVD. Quality Views were analyzed for the same shoebox spaces via a grasshopper-based algorithm, developed from the LEED v4 evaluation framework for Quality Views. The dataset was further analyzed with an Artificial Neural Network algorithm written in Python. The accuracy of the predictions was estimated at 97% on average. The developed model could be used in early design stages analyses without the need for time-consuming simulations in previously used platforms and programs.

## Key Innovations

- A machine learning framework for early design stages analysis of daylight, glare, and quality views altogether

## Practical Implications

- Application of artificial neural networks in predicting visual comfort performance of a space

## Introduction

Due to binding effect on occupant wellbeing and energy savings, Daylight and visual comfort assessment, have been within the comprehensive discussion of sustainable building design in recent years (Mariana G. Figueiro, 2017). Besides, lighting energy has previously consumed almost 19% of total electricity consumption (IEA, 2006), and office buildings consume more than 20% of this energy (Ürge-Vorsatz, 2012). Nowadays 7% of consumed energy in buildings is for lighting; thanks to the progress in daylight standards and regulations (IEA, 2019). Besides, lighting energy represents a main source of heat that directly impacts the building's cooling energy consumption (Tzempelikos & K.Athienitis, 2007). Accordingly, Proper daylighting strategies seriously affects the optimiziation of the lighting and cooling energies, especially in harsh climates (Ayoub, 2019). One of the sustainable procedures to enhance the mentioned efficiency is to use daylighting strategies that provide the controlled use of natural daylight inside buildings (Reinhart, 2014). Quantification of internal daylighting conditions have been done via simple methods, protractors, scale models, mathematical formulas and rules-of-thumb in the early years of daylight predictions (Ayoub, 2020). In recent years, because of the progress of computer processing, newly introduced methods, named white-box, extended the former approaches (Ward, 1994). constant efforts have been made to integrate all available techniques into design process, and Climate-based daylight metrics are now widely used as the most accurate indicators of the overall daylight situation of the spaces (Shaghaghian & Yan, 2020); However, The tools and applied professional methods that predict daylighting performance in buildings are still impractical (Emilie Nault, 2017).Time-intensive CBDM can be a real challenge when assessing the daylight performance for numerous design alternatives, especially in parametric design environments when designers aim to push for daylight optimization (C. L. Lorenza, 2019). Besides architects don't spend time to include what they consider complex methods into their practices (Mardaljevic, 2015). Artificial intelligence applications are achieving significant successes in many fields today. Predictive models that are built on machine learning algorithms, called black-box, have been receiving recognition from the building design community because of their ability to handle complex problems in a short time with acceptable accuracy. These newly introduced methods are based on algorithms that learn the mathematical relationship between a dataset's parameters and construct a mathematically-fit model. Much useful information about the data could then be extracted without the need for time-consuming simulations or calculations.The application of these newly introduced methods in daylighting prediction is still underexploited (Ayoub, 2020).

Studies on the application of MLAs in daylighting has been progressive in recent years. The objectives of these studies are varied, including building type, location, selected algorithm, and output parameters.

Da Fonseca et al. in 2013, assessed the prediction quality of two known algorithms named artificial neural networks and multiple linear regression in a potential energy saving of a building through daylighting. They figured out that even though the ANN showed better results, both algorithms could predict the results with high accuracy.

In similar research, Chatzikonstantinou and Sariyildiz in 2015, evaluated three algorithms, including ANN, K-nearest, and random forest. The research is focused on two metrics: daylight autonomy and daylight glare probability. The results accordingly indicate that machine-learning-based approaches achieve a favorable trade-off between accuracy and computational cost. The authors suggest that they provide a worthwhile alternative for performance evaluations during architectural conceptual design.

Mohammad Ayoub, in 2020, has reviewed previous researches in this field. The author studied related articles in the following categories: Scope of prediction, MLAs, Data sources, sizes and temporal granularities, and evaluation metrics. According to the results, the most used algorithm in these studies are artificial neural networks, multiple linear regression and support vector machine, considering regression as the most mentioned problem to solve. And finally, the most used output parameters are illuminance values, daylight autonomy, spatial daylight autonomy, and daylight glare probability, respectively.

In this study, a framework will be formed based on Artificial Neural Network algorithm to predict daylight and visual comfort metrics in a simple room. The main purpose is to evaluate the algorithm in the estimation of the mentioned metrics and determine whether this method could replace former tools and software.

For this purpose, a dataset is primarily developed based on simulations derived results from Honeybee for Grasshopper. The dataset includes daylight, glare, and quality view metrics. The dataset is then analyzed with the machine learning algorithm presented. The significant feature of this research is that three fields of visual comfort assessment are analysed simultaneously, which gives a complete vision about the intended space.

## Methodology

To assess MLAs in predicting the intended metrics, two main steps were taken. At first, a dataset was primarily created from simulation-derived results that emerged in grasshopper for rhino and the related plugins. Then the dataset is analysed in an MLA, and the predictive model is created. The details of these steps are as below:

### Dataset creation

The defined space was a one-side-opened shoebox room that several parameters of it were considered as variables to define the alternatives; they include window orientation, space dimensions, interior surfaces reflectance factor, shading states, O.K.B, window height, and window divisions. The shape of the space is presented in fig.1. the variables interval changes could be found in table 1.

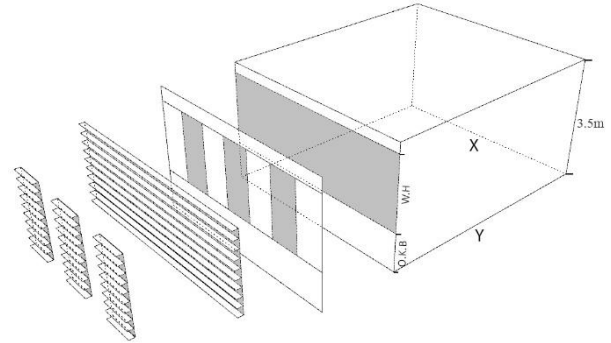

*Figure 1.The shoebox space form*

*Table 1. variables interval changes*

| | Variables | | Number of alternatives |
|---|---|---|---|
| Fixed Parameters | Location | Tehran | - |
| | Occupation | Office | - |
| | Window glass type | Single pane glass with 0.85 visual transmittance | - |
| | Space height(m) | 3.5 | - |
| Variables | Window orientation | N,E,S,W | 4 |
| | Space dimensions(x,y) (m) | (3,4)-(6,7)-(8,10) | 3 |
| | interior surfaces reflectance factor | 0.2 - 0.4 – 0.7 | 3 |
| | Shading state | No shading – 15cm horizontal louvre | 2 |
| | Window Sill Height (m) | 0.5 – 0.7 – 0.9 – 1.1 | 4 |
| | Window height (m) | 1.2 – 1.5 – 1.8 – 2.1 – 2.4 | 5 |
| | window divisions | 1 window as wide as the rooms width- 3 windows with equal distances | 2 |
| | Overall alternatives | | 2880 |

To select appropriate metrics for assessment of the space, it was considered that the chosen metrics should indicate the overall performance of the space in the fields of daylight sufficiency, glare and visual comfort, and quality views. The selected metrics in daylight and glare analyzes include: mean daylight autonomy, spatial daylight autonomy, useful daylight illuminance, annual sunlit exposure, and spatial visual discomfort. The threshold considered for minimum acceptable illuminance of the space is 300lux. Quality views are assessed by the terms of LEED v4 principles. This standard states four main factors for evaluation of views in the buildings: view factor, view depth, view range, and the characteristics of the landscape outside the window. If two out of these four terms are provided for at least 75% of the space, the view quality of the room is acceptable. Due to the definition of assumed variables in this project, the landscape outside the window is not mentioned and assessed, so three other factors are simulated and discussed. The modeling process of the shoebox space was done in grasshopper for rhino. Honeybee v0.0.66 was applied as the simulation engine in daylight and glare metrics. The quality view factors were geometrically calculated in grasshopper. (fig.3) The Daysim simulation engine factors were adjusted in accordance with table 2.

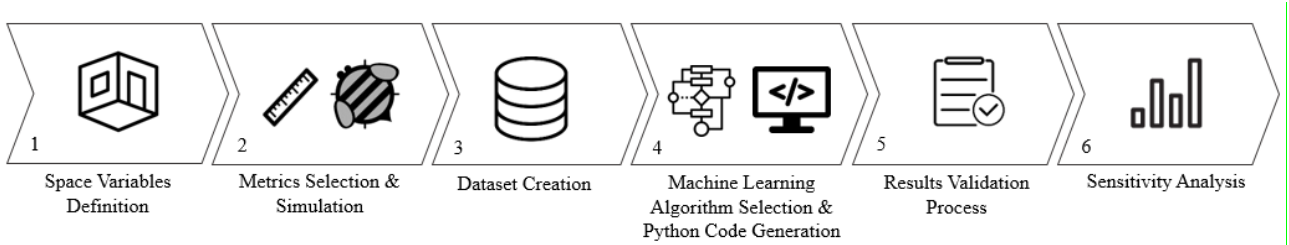

Figure 2. Overall Research Methodology Steps

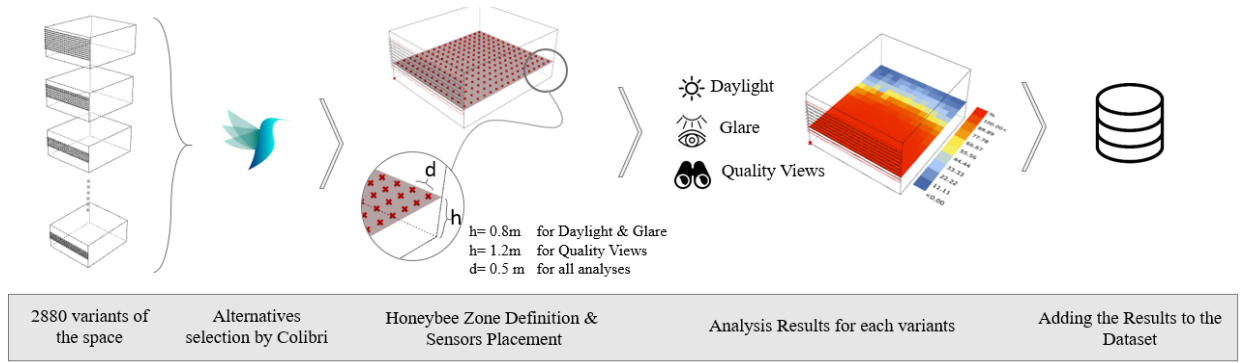

Figure 3. grasshopper code for daylight, glare and quality views calculations schematic overview

Table 2. daysim parameters adjustment

| Radiance parameters | value |
|---|---|
| ab | 5 |
| ad | 1500 |
| as | 128 (default) |
| ar | 16 (default) |
| aa | 0.25 (default) |

**Machine learning algorithm development**

As mentioned above, several algorithms have been used to predict daylight metrics in recent studies. The most common algorithm is artificial neural networks with acceptable prediction accuracy. (Ayoub, 2020) In this research, this algorithm was applied for producing the predictive model.

Artificial neural networks is an algorithm that does basically non-linear function approximations and performs numerical predictions via supervised, unsupervised and reinforcement learning techniques (Yao, 1999) the biological structure of neural networks that constitutes the brain and nervous system is the main concept behind ANN. This algorithm is composed of a several layers, each of which contains several processing elements, called neurons. The input layer receives the input data; the hidden layers perform the computations, and output layer predicts the results. (Hecht-Nielsen, 1992) Inputs are multiplied by corresponding weights, and the sum is then biased to minimize errors that stem from the difference between the actual and predicted outputs. The value passes through an activation function, acting as a gate that allows the transmittance of the value to the next layer. (Bishop, 2006) There are key parameters that tune ANN; they include number of hidden layers, number of hidden neurons, learning rate, number of epochs, batch size, and activation functions. These parameters are selected and tuned for each model and differs from one dataset to another. The hyperparameters of the ANN are presented in table 3. The algorithm architecture has a single hidden layer with 40 neurons.

---

[1] Ambient Bounces
[2] Ambient Divisions
[3] Ambient Super-Samples
[4] Ambient Resolution
[5] Ambient Accuracy

This architecture had the best performance among all alternatives. Tuning the algorithm included several try and error steps that was taken in order to get the best performing algorithm architecture. This process started with a 3-layer network with varied number of neurons from 10 to 40. At each step, these parameters were adjusted and the results were observed. It was noticed that the algorithm with a single hidden layer and 40 neurons performed better compared to others. After adjusting the hidden layer properties, epoch and batch sizes were adjusted through another try and error process. The influence of the changes in these two latter parameters are not of vital significance. So that, a simple and fast responding value was finally chosen for them.

It should be noted that the window orientation variable has been divided into 4 inputs with 0 and 1 values e.g., north orientation is defined as 4 inputs like [1,0,0,0] for the neural network model. This was done because if the four orientations were defined as 1 to 4 numerical values, each orientation would be weighed compared to another.

*Table 3. hyper parameters of the best performing ANN algorithm*

| ANN hyper parameters | values |
|---|---|
| Train/ test data numbers | 2304/576 (80% to 20%) |
| Number of layers | 1 |
| Number of neurons | 40 |
| Epoch | 50 |
| Batch size | 10 |

To assess the accuracy of the prediction results of different algorithm architectures, two error metrics were used: Mean square error and mean actual error.

MSE defines the average actual difference between predicted and simulated values in the dataset, while MSE is the average squared difference between the estimated values and the actual value. MAE fails to punish large errors in prediction, while MSE indicates large errors more obviously.

$$MAE = \frac{\sum|y_i - \hat{y}_i|}{n}$$

$$MSE = \frac{\sum(y_i - \hat{y}_i)^2}{n}$$

By comparing the predicted and simulated results of a new set of alternatives for the shoebox space, the ANN-based model's efficiency was evaluated. The new dataset included new values for the shoebox space parameters that didn't exist in the former dataset. This was mainly to evaluate the machine learning model performance confronting totally new inputs that the ANN didn't learn from in the training step. The alternatives considered for this step are presented in table 4.

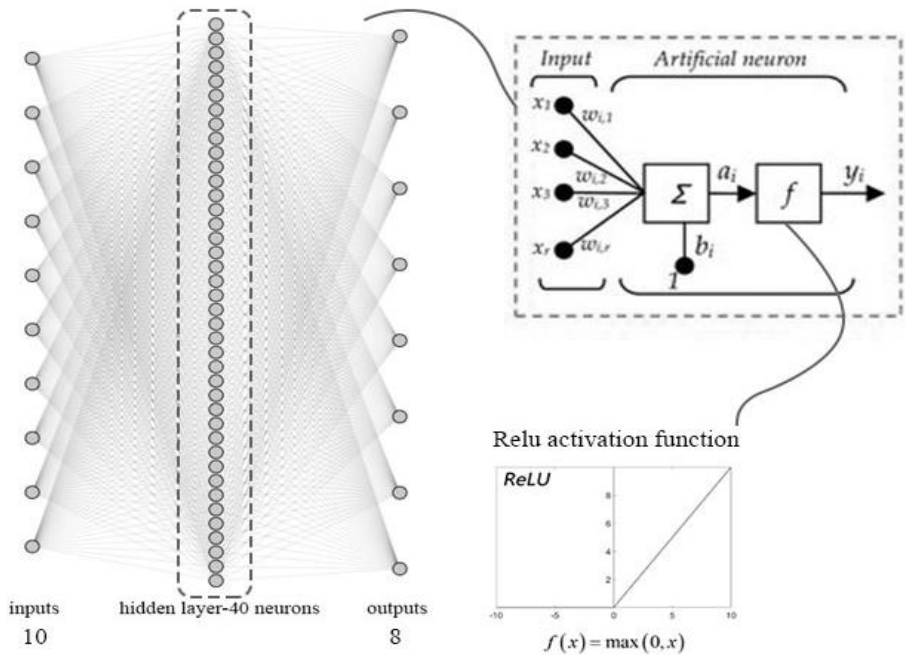

*Figure 4. Applied artificial neural network architecture*

*Table 4. variables interval changes for validation step*

|  | Variables |  | Number of alternatives |
|---|---|---|---|
| Fixed Parameters | Location | Tehran | - |
|  | Occupation | Office | - |
|  | Window glass type | Single pane glass with 0.85 visual transmittance | - |
|  | Space height(m) | 3.5 | - |
| Variables | Window orientation | S,E | 2 |
|  | Space dimensions(x,y) (m) | (7,8)-(5,6) | 2 |
|  | interior surfaces reflectance factor | 0.3-0.6 | 2 |
|  | Shading state | No shading – 15cm horizontal louvre | 2 |
|  | Window O.K.B (m) | 0.8 - 1 | 2 |
|  | Window height (m) | 1.6 – 2.0 | 2 |
|  | window divisions | 1 window as wide as the rooms width | 1 |
|  | Overall alternatives |  | 64 |

**Sensitivity Analysis with Shapley additive explanations**

There are several procedures for sensitivity analysis of the metrics to different variables. One of the most used methods is the Pierson factor calculation. This factor is used primarily for the values which have a linear relationship which each other (Magiya, 2019). In our case in this research, there's no confidence about the relationship among the values. That's because the ANN-based model is categorized as a black-box model meaning that the exact function and formula calculating the metrics is not apparent (Ayoub, 2020). Accordingly, another method named Shapley additive explanations, is applied for the sensitivity analysis. The main aspect of this method is that it doesn't necessarily adopt linear functions for the analysis (Rathi, 2020) (Lundberg, et al., 2017). In this method, the algorithm performance is evaluated in two different conditions, with and without the target variable's presence. Then each feature's effect on the prediction process would be observed by a value named Shap value. Shap value is calculated through this equation:

$$\Phi_i = \sum_{S \subseteq M \setminus i} \frac{|S|!\,(|M|-|S|-1)!}{|M|!} [f(S \cup i) - (S)]$$

The main result of this equation is the difference between the model's prediction with and without feature i. S is a subset of the features, that doesn't include the feature that the Φ value is calculated for. $S \cup i$ is a subset of features that include S features and i feature together $S \subseteq M$ includes all S sets that are subsets of M perfect set excluding i.

## Results and Discussion

**The neural networks validation results**

All 64 alternatives forming the validation dataset calculated by both the ANN model and CBDM simulations were compared to figure out the efficiency of the MLA-based method. The results were promising in nearly all eight metrics that were analyzed. (figure 6.) Both MAE and MSE error metrics were used again at this step to assess the prediction accuracy of the ANN model (table 5). The results indicate that on average, daylight metrics had more promising results than the others. However, all errors are within the acceptable range, and the model's prediction accuracy is satisfying considering the fact that there would be no need for time-consuming simulations and the overall performance of the space in daylight, glare, and quality views would be available in a short time.

Each variable has a distinct effect on the calculated metrics, which is indicated with the Shap value. The application of this analysis could be the further usage in extending the model in variables for prediction of more varied alternatives; e.g., if window height is the most influential variable, the extension of the model could be formed based on more number of variations in this value.

*Table 5. Metrics prediction errors*

|  |  | MAE | MSE |
|---|---|---|---|
| **Daylight** | UDI | 0.022 | 0.0008 |
|  | mDA | 0.019 | 0.0006 |
|  | sDA | 0.042 | 0.003 |
| **Glare** | sVD | 0.047 | 0.003 |
|  | ASE | 0.031 | 0.003 |
| **Quality Views** | View Range>90° | 0.06 | 0.007 |
|  | View Depth | 0.018 | 0.0017 |
|  | View Factor | 0.03 | 0.0015 |

The results which are presented in figure 6 and table 5 indicate the results in the view range factor has the most MAE value in comparison to the other factors. The errors in this metric vary from 0.01 to 0.11 in the actual value of the metric. This could be a significant error when it comes to the idea of replacing this method with previous calculation procedures; but it must be noted that these results are emerged from a limited set of data that most of its variables may have no effect on quality view metrics. These variables include interior surfaces reflectance, window orientation and shading states.

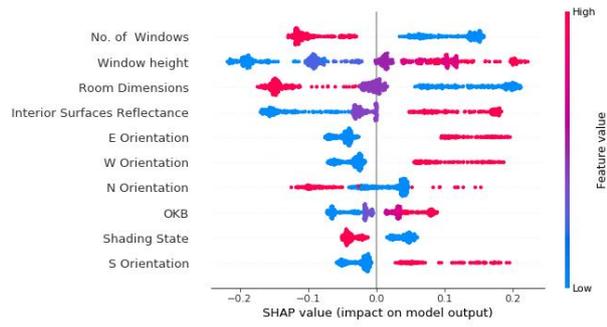
mDA

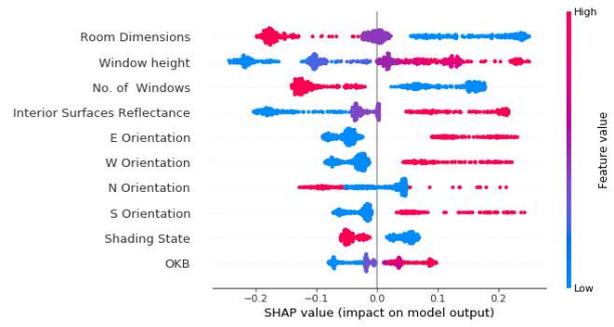
SDA

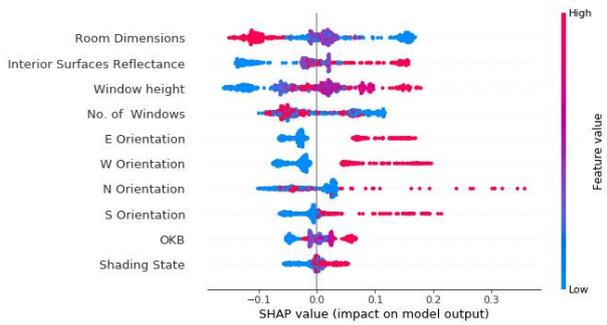
UDI

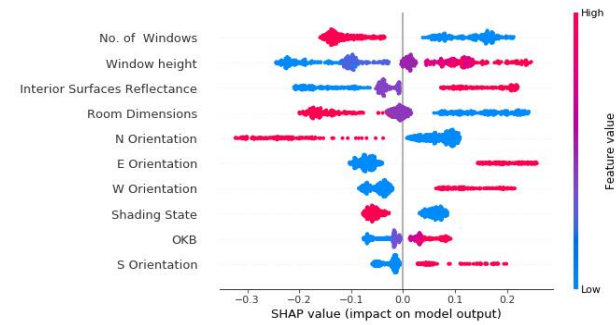
ASE

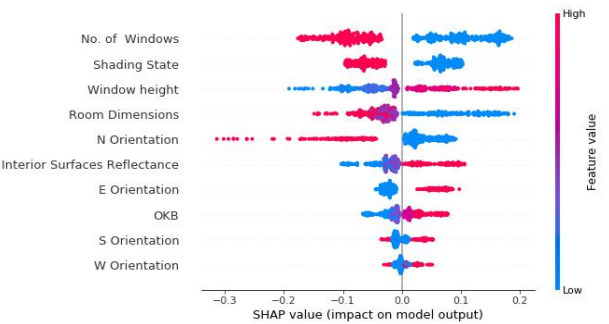
sVD

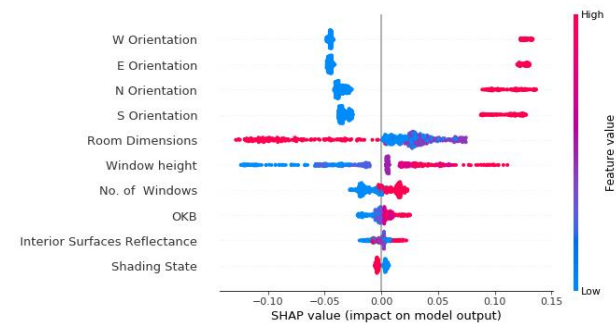
View Factor

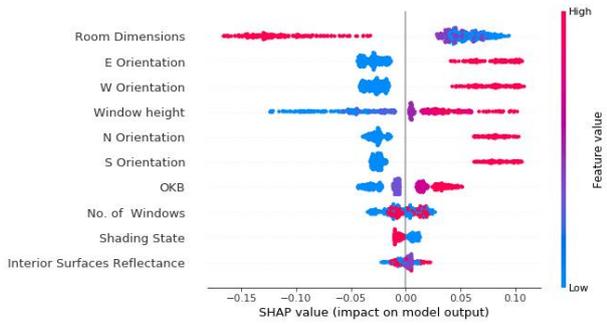
View Depth

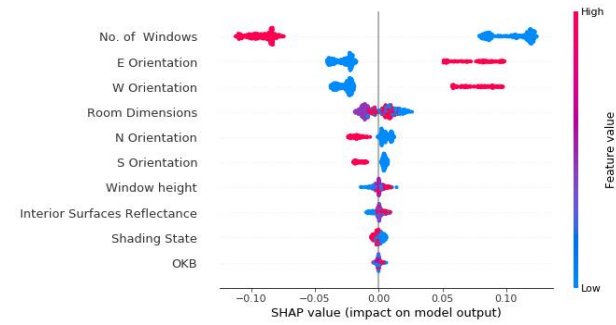
View Range

*Figure 5. Shap values (Impact on model output) for each variable on each metric*

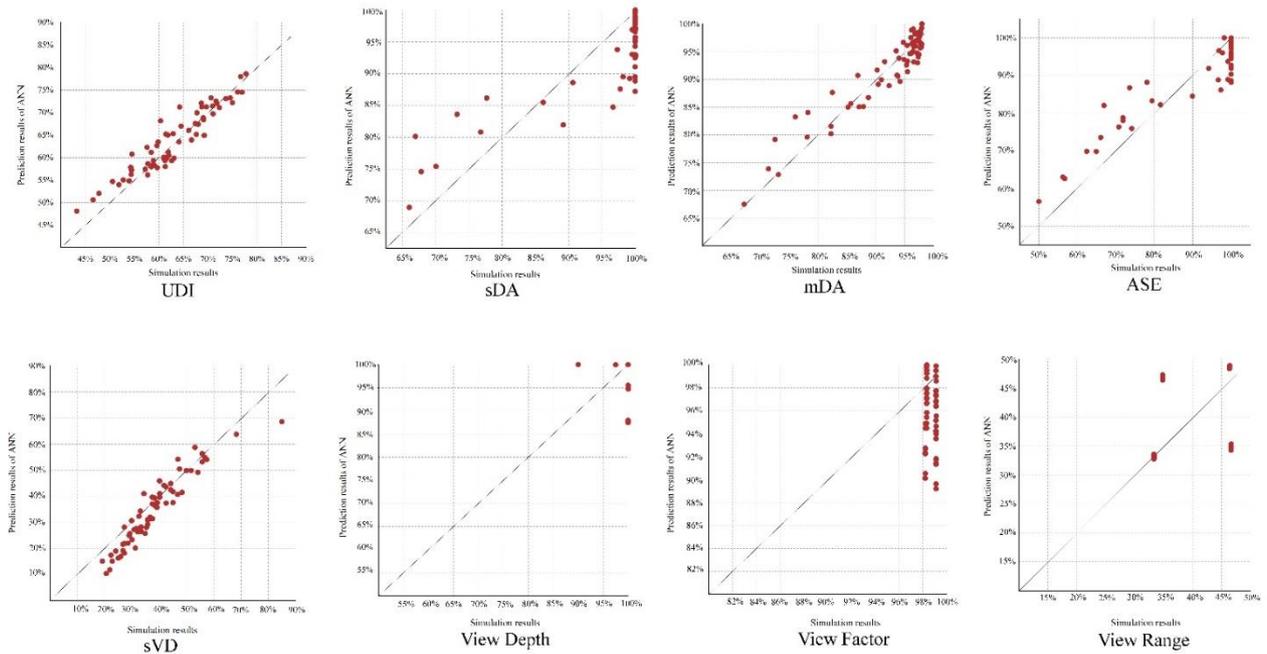

*Figure 6. Simulated vs. Predicted values of the metrics*

Accordingly, overall 2880 data will not be applicable to the models training process and the model won't be able to conduct the predictions as well as other daylight and visual comfort metrics. It is recommended to apply more variables regarding physical parameters of the analyzed space to get more promising results in quality view metrics as well. Other metrics which indicate daylight sufficiency and glare may have overall performance with same MAE and MSE values but actual errors on each alternative are hardly more than 10%.

The predictive model which was formed in this research may not be practically used instead of traditional methods for now, but the results are to be applied in performance simulation analyses in two main fields. Firstly, all three fields of study in daylight and visual comfort are simultaneously assessed in the study and the results indicate that these new methods could be applied for all mentioned fields. Secondly, by sensitivity analyses we could be informed of the most effective variables which could be extended for further researches which are focused on improvements of this field of study.

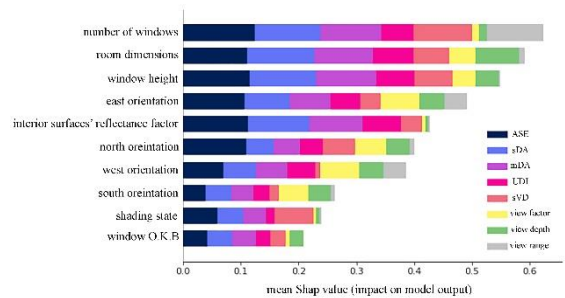

*Figure 7. Sensitivity analysis of the metrics*

The sensitivity analysis results show that the number of windows, room dimensions, and window height are the most effective variables in the set. In contrast, the sill height and shading state are the least effective ones. Figure 5 indicates the Shap value analysis for each of the metrics. Some of the variables have negative effect on the output parameter and others may have positive ones. In the graphs presented in figure 5, the plotted colors indicate the effect of each variable value in the dataset on the output parameter. For instance, The effect of orientation on the view range metric is significantly high. While it seems this results could be illogical, it shall be noted that orientation is one of the affective factors on the width of each alternative's window in our spaces; so that, these results are exclusively related to the present dataset and cannot be extended to other situations or space definitions. The average influence is represented in figure 7.

## Conclusions

An ANN-based predictive model has been developed in this research. Daylight and visual comfort, along with quality views metrics, have been analysed in a single shoebox space. The results indicated that all three fields of metrics could be predicted by this method with acceptable accuracy. This means that this procedure could form a framework to be substituted with the present simulation-based methods, and the calculation of the metrics would be available without the need for time-consuming computer-based simulations.

To expand the created predictive model of this research, the number of variables and their interval changes shall be extended to new values in the firstly simulated dataset. This will lead to more accurate predictions and more reliable results based on proper training of the ANN model.

## References

Ayoub, M. (2019). A multivariate regression to predict daylighting and energy consumption of residential buildings within hybrid settlements in hot-desert climates. *Indoor and built environment, 28*, 848-866.

Ayoub, M. (2020). A review on machine learning algorithms to predict daylighting inside buildings. *Solar energy, 202*, 249-275.

Bishop, S. (2006). *Pattern Recognition and Machine Learning.* Singapore: Springer.

C. L. Lorenza, A. B. (2019). Artificial Neural Networks for parametric daylight design. *Architectural Science and Review, 63*, 210-221.

Chatzikonstantinou, I., & Sariyildiz, S. (2015). Approximation of simulation-derived visual comfort indicators in office spaces: a comparative study in machine learning. *Architectural Science Review* , 1-16.

Emilie Nault, P. M. (2017). Predictive models for assessing the passive solar and daylight potential of neighborhood designs: A comparative proof-of-concept study. *Building and Environment, 116*, 1-16.

Hecht-Nielsen, R. (1992). Theory of the backpropagation neural network. In *Neural networks for perception* (pp. 65-93). Academic press.

IEA. (2006, Jun 26). *Light's Labour's Lost" – Policies for Energy-efficient Lighting,International Energy Agency*. (International Energy Agency) Retrieved from https://www.iea.org/news/lights-labours-lost-policies-for-energy-efficient-lighting

IEA. (2019). *Energy Efficiency,International energy agency*. (International energy agency) Retrieved from https://www.iea.org/reports/energy-efficiency-2019

Lundberg, M., S., & Lee, S.-I. (2017). A unified approach to interpreting model predictions.

Magiya, J. (2019). *Pearson Coefficient of Correlation Explained.* Retrieved from towardsdatascience.com: https://towardsdatascience.com/pearson-coefficient-of-correlation-explained-369991d93404

Mardaljevic, J. (2015). Climate-based daylight modelling and its discontents. *Simple Buildings Better Buildings? Delivering Performance through Engineered Solutions, CIBSE Technical Symposium.* London.

Mariana G. Figueiro, B. S. (2017). The impact of daytime light exposures on sleep and mood in office workers. *Sleep Health, 3*(3). Retrieved from https://doi.org/10.1016/j.sleh.2017.03.005

Rathi, P. (2020). *A Novel Approach to Feature Importance — Shapley Additive Explanations.* Retrieved from towardsdatascience: https://towardsdatascience.com/a-novel-approach-to-feature-importance-shapley-additive-explanations-d18af30fc21b

Reinhart. (2014). *Daylighting handbook I: Funamentals, designing with the sun, Muscle & Nerve.* . Building Technology Press.

Shaghaghian, Z., & Yan, W. (2020). Application of deep learning in generating desired design options: Experiments using synthetic training dataset. *2020 Building Performance Analysis Conference and SimBuild co-organized by ASHRAE and IBPSA-USA*.

Tzempelikos, A., & K.Athienitis, A. (2007). The impact of shading design and control on building cooling and lighting demand. *Solar Energy, 81*, 369-382.

Ürge-Vorsatz, D. E. (2012). *Global Energy Assessment, Toward a sustainable future.* Cambridge University Press.

Ward, G. (1994). The RADIANCE lighting simulation and rendering system. *Proceedings of the 21st annual conference on Computer Graphics and Interactive Techniques.*

Yao, X. (1999). Evolving artificial neural networks. IEEE- Institute of electrical and electronics engineers.